\documentclass[10pt,twocolumn,letterpaper]{article}

\usepackage{cvpr}              

\usepackage[dvipsnames]{xcolor}

\newcommand{\beq}{\vspace{0mm}\begin{equation}}
\newcommand{\eeq}{\vspace{0mm}\end{equation}}
\newcommand{\beqs}{\vspace{0mm}\begin{eqnarray}}
\newcommand{\eeqs}{\vspace{0mm}\end{eqnarray}}
\newcommand{\barr}{\begin{array}}
\newcommand{\earr}{\end{array}}

\newcommand{\R}{\mathbb{R}}

\usepackage{color, colortbl}
\definecolor{Gray}{gray}{0.93}

\usepackage{xcolor,amsmath}
\usepackage[linesnumbered,ruled,vlined]{algorithm2e}
\DontPrintSemicolon

\SetKwComment{Comment}{\color{green!50!black}\# }{}

\SetKwProg{Function}{def}{:}{}

\SetKwProg{For}{for}{:}{}
\SetKwProg{If}{if}{:}{}

\definecolor{cvprblue}{rgb}{0.21,0.49,0.74}
\definecolor{myYellow}{RGB}{227, 227, 0}
\usepackage[pagebackref,breaklinks,colorlinks,linkcolor=cvprblue,urlcolor=cvprblue,citecolor=cvprblue]{hyperref}
\usepackage{multirow}
\usepackage{colortbl}
\usepackage{xcolor}

\def\paperID{16452} 
\def\confName{CVPR}
\def\confYear{2024}

\title{VideoGrounding-DINO: Towards Open-Vocabulary Spatio-Temporal Video Grounding}

\author{
  Syed Talal Wasim$^{1}$ \quad 
  Muzammal Naseer$^{1}$ \\ 
  Salman Khan$^{1,2}$ \quad
  Ming-Hsuan Yang$^{3, 4}$ \quad
  Fahad Shahbaz Khan$^{1,5}$
  \vspace{0.5em} \\
  $^{1}$Mohamed bin Zayed University of AI \quad 
  $^{2}$Australian National University \\ 
  $^{3}$University of California, Merced \quad 
  $^{4}$Google Research \quad
  $^{5}$Link\"{o}ping University \quad
}

\begin{document}
\maketitle
\begin{abstract}
Video grounding aims to localize a spatio-temporal section in a video corresponding to an input text query. This paper addresses a critical limitation in current video grounding methodologies by introducing an Open-Vocabulary Spatio-Temporal Video Grounding task. Unlike prevalent closed-set approaches that struggle with open-vocabulary scenarios due to limited training data and predefined vocabularies, our model leverages pre-trained representations from foundational spatial grounding models. This empowers it to effectively bridge the semantic gap between natural language and diverse visual content, achieving strong performance in closed-set and open-vocabulary settings. Our contributions include a novel spatio-temporal video grounding model, surpassing state-of-the-art results in closed-set evaluations on multiple datasets and demonstrating superior performance in open-vocabulary scenarios. 
Notably, the proposed model outperforms state-of-the-art methods in closed-set settings on VidSTG (Declarative and Interrogative) and HC-STVG (V1 and V2) datasets. Furthermore, in open-vocabulary evaluations on HC-STVG V1 and YouCook-Interactions, our model surpasses the recent best-performing models by 4.88 m\_vIoU and 1.83\% accuracy, demonstrating its efficacy in handling diverse linguistic and visual concepts for improved video understanding. Our codes will be publicly released.
\end{abstract}    
\section{Introduction}
\label{sec:intro}

\begin{figure}[t]
\centering
    \includegraphics[width=\columnwidth]{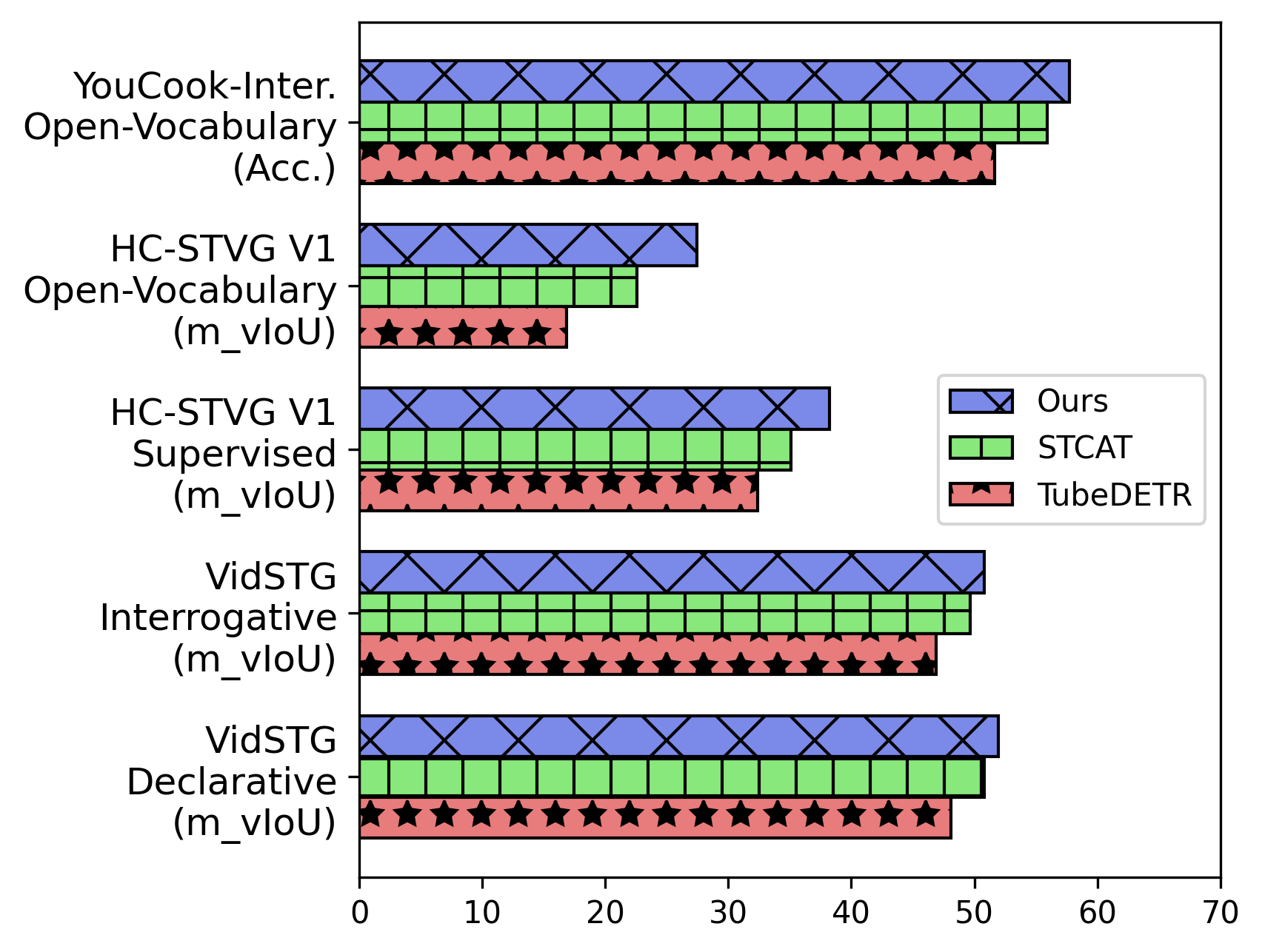}
    \caption{\textbf{Performance comparison} on conventional closed-set and open-vocabulary settings for the video grounding task. We compare our approach with TubeDETR~\cite{yang2022tubedetr} and STCAT~\cite{jin2022stcat} in supervised setting for VidSTG~\cite{zhang2020stgrnVIDSTG} declarative/interrogative and HC-STVG V1~\cite{tang2021stgvtHCSTVG}, along with open-vocabulary evaluation on HC-STVG V1 and YouCook-Interactions~\cite{tan2021youcook-inter} datasets.}
    \label{fig:intro_fig}
\end{figure}

Spatio-temporal video grounding is pivotal in linking visual content with natural language descriptions, thus facilitating semantics interpretation within visual data.
Prevailing approaches in video grounding such as TubeDETR~\cite{yang2022tubedetr}, STCAT~\cite{jin2022stcat}, and STVGFormer~\cite{lin2023stvgformer}  focus mainly on supervised closed-set settings, where models are trained on specific datasets~\cite{zhang2020stgrnVIDSTG, tang2021stgvtHCSTVG} with predefined vocabulary and meticulously annotated data. While these current state-of-the-art methods excel in closed-set settings on datasets like VidSTG~\cite{zhang2020stgrnVIDSTG} and HC-STVG~\cite{tang2021stgvtHCSTVG}, their limited generalization beyond training dataset distributions poses a significant challenge. The relatively small scale and restricted sample variety in existing video datasets hinder models from adapting to unseen scenarios effectively.

Motivated by the inherent limitation of supervised closed-set approaches in terms of their restricted vocabulary, this paper investigates Open-Vocabulary Spatio-Temporal Video Grounding. Unlike conventional methodologies, this paradigm addresses challenges posed by the unrestricted diversity of language and visual concepts in the wild. The goal is to train on a set of base categories and generalize to unseen objects/actions based on the open-vocabulary nature of backbone models. This paper explores the challenges and opportunities inherent in open-vocabulary video grounding, laying the groundwork for more robust and versatile video understanding.

However, training an effective open-vocabulary video grounding model would require a large enough dataset with a rich set of natural language expressions and corresponding spatio-temporal localizations. Such an extensive dataset can allow the model to learn generalized visual and textual representations and handle out-of-distribution samples. However, video grounding datasets \cite{zhang2020stgrnVIDSTG, tang2021stgvtHCSTVG} are quite limited in scale, e.g., VidSTG has only 5.4k training videos with 80.6k distinct sentences. In contrast, an image grounding model GLIP \cite{li2021glip} is trained with $\sim$26.5M image-text pairs. Therefore, our research explores the following fundamental question: \emph{How spatio-temporal video grounding models can achieve robust performance in both Closed-Set and Open-Vocabulary scenarios without requiring large-scale video annotations, ensuring effective generalization beyond training datasets?}

In addressing this question, we find inspiration in the accomplishments of foundational models specializing in spatial grounding~\cite{huang2023kosmos-1, peng2023kosmos-2, chen2023shikra, you2023ferret, liu2023gdino}. These models undergo training on an extensive corpus of image-text data, enabling effective generalization to samples from a given target distribution. We aim to harness this pretrained representation to enhance our video grounding model. Our proposed solution is a spatio-temporal video grounding model adopting a DETR-like architecture enhanced by temporal aggregation modules. The spatial modules are initialized using the pretrained representation of a foundational image model \cite{liu2023gdino}. Meanwhile, the image and text feature extractors remain frozen while the video-specific spatio-temporal adaptations are modeled via learnable adapter blocks. This approach is designed to preserve the nuanced representation of the foundational model, enhancing our model's ability to generalize effectively to novel samples.

A summary of our closed-set and open-vocabulary results is shown in \autoref{fig:intro_fig}, where the proposed approach excels in both settings by a clear margin. Our major contributions are summarized as follows:
\begin{itemize}\setlength{\itemsep}{0em}
    \item For the first time, we evaluate spatio-temporal video grounding models in an open-vocabulary setting on HC-STVG V1~\cite{tang2021stgvtHCSTVG} and YouCook-Interactions~\cite{tan2021youcook-inter} benchmarks in a zero-shot manner. We outperform state-of-the-art methods TubeDETR~\cite{yang2022tubedetr} and STCAT~\cite{jin2022stcat} by $4.26$ m\_vIoU and $1.83 \%$ accuracy, respectively.
    \item By combining the strengths of spatial grounding models with complementary video-specific adapters, our approach consistently outperforms the previous state-of-the-art in closed-set setting on four benchmarks, i.e., VidSTG (Declarative)~\cite{zhang2020stgrnVIDSTG}, VidSTG (Interrogative)~\cite{zhang2020stgrnVIDSTG}, HC-STVG V1~\cite{tang2021stgvtHCSTVG} and HC-STVG V2~\cite{tang2021stgvtHCSTVG}.
\end{itemize}
\section{Related Work}
\label{sec:related}

\noindent \textbf{Spatial Grounding Foundation Models:} Recent literature introduces notable spatial grounding models. 
GLIP~\cite{li2021glip} unifies object detection and phrase grounding through a language-image pre-training model, leveraging extensive image-text pairs for semantic-rich representations. 
Grounding DINO~\cite{liu2023gdino} integrates language with a transformer-based detector to achieve an open-set detector, excelling in benchmarks like COCO and ODinW. 
Kosmos-1~\cite{huang2023kosmos-1} and Kosmos-2~\cite{peng2023kosmos-2} contribute Multimodal Large Language Models (MLLMs) with capabilities such as zero-shot and few-shot learning, language understanding, and multimodal tasks. Kosmos-2~\cite{peng2023kosmos-2} specifically integrates grounding into downstream applications, introducing GrIT, a large-scale dataset of Grounded Image-Text pairs. Shikra~\cite{chen2023shikra} addresses referential ability in MLLMs by handling spatial coordinates in inputs and outputs, showcasing promising performance in various vision-language tasks. Ferret~\cite{you2023ferret} unifies referring and grounding in the LLM paradigm, achieving superior performance in classical referring and grounding tasks, excelling in region-based multimodal chatting and image description. Recently, GLaMM~\cite{hanoona2023GLaMM} allows pixel-level grounded conversations with an LLM, showcasing generalizability to several captioning and referring segmentation tasks. However, spatial methods cannot work for grounding objects in videos, a gap addressed by this work.

\noindent \textbf{Spatio-Temporal Video Grounding:} 
Several methods tackle the challenge of localizing objects in untrimmed videos based on query sentences. STVGBert~\cite{su2021stvgbert} presents a one-stage visual-linguistic transformer for simultaneous spatial and temporal localization. 
TubeDETR~\cite{yang2022tubedetr} introduces a transformer-based architecture to model temporal, spatial, and multi-modal interactions efficiently. 
Augmented 2D-TAN~\cite{tan2021aug2d} adopts a two-stage approach, enhancing the 2D-TAN with a temporal context-aware Bi-LSTM Aggregation Module. 
OMRN~\cite{zhang2022omrn} addresses the challenge of unaligned data and multi-form sentences in spatio-temporal video grounding, proposing an object-aware multi-branch relation network for effective relation discovery. 
MMN~\cite{wang2022mmn} introduces a Mutual Matching Network as a metric-learning framework for temporal grounding, achieving competitive performance.
STCAT~\cite{jin2022stcat} is an end-to-end one-stage framework addressing feature alignment and prediction inconsistency.  
Finally, STVGFormer~\cite{lin2023stvgformer} proposes an effective framework with static and dynamic branches for cross-modal understanding. 
While the above methods advance video grounding, their generalization to out-of-distribution and open-vocabulary samples is limited due to constrained video datasets~\cite{zhang2020stgrnVIDSTG, tang2021stgvtHCSTVG}. To address this issue, we utilize the generalized representation of spatial grounding foundation models~\cite{huang2023kosmos-1, peng2023kosmos-2, chen2023shikra, you2023ferret, liu2023gdino, li2021glip}  trained on a large corpus of image-text data 
and can perform well on both closed-set and open-vocabulary evaluations.
\section{Methodology}
\label{sec:method}

As discussed above, the current state-of-the-art spatio-temporal video grounding methods~\cite{zhang2022omrn, jin2022stcat, lin2023stvgformer, su2021stvgbert, yang2022tubedetr, tan2021aug2d, tang2021stgvtHCSTVG, wang2022mmn, yu2022yu2rd, zhang2020stgrnVIDSTG} primarily evaluate in a supervised setting on the VidSTG~\cite{zhang2020stgrnVIDSTG} and HC-STVG~\cite{tang2021stgvtHCSTVG} datasets. However, these methods lack the multimodal spatio-temporal understanding required to perform well on out-of-distribution samples~\cite{chen2023what}. Therefore, this work aims to achieve improved \emph{open-vocabulary} performance while maintaining strong \emph{closed-set} video-grounding performance.

We take inspiration from recent foundation models for spatial grounding~\cite{huang2023kosmos-1, peng2023kosmos-2, you2023ferret, chen2023shikra, li2021glip, liu2023gdino}. These models are trained on a large corpus of visual-textual data and hence, generalize well to unseen samples. We aim to leverage the strong generalization capabilities of such foundation models to achieve strong open-set spatio-temporal video grounding performance. Our proposed spatio-temporal video grounding method uses DETR-like~\cite{carion2020detr} design, with temporal aggregation and adaptation modules for learning video-specific representations.

Below, we explain our proposed methodology. We formally define the spatio-temporal video grounding problem in \autoref{sec:meth:problem}. We then explain our architecture details in \autoref{sec:meth:video_grounding}. We finally explain the loss formulation used to train the model and model initialization in \autoref{sec:meth:loss}.

\begin{figure*}[t]
\centering
    \includegraphics[width=\textwidth]{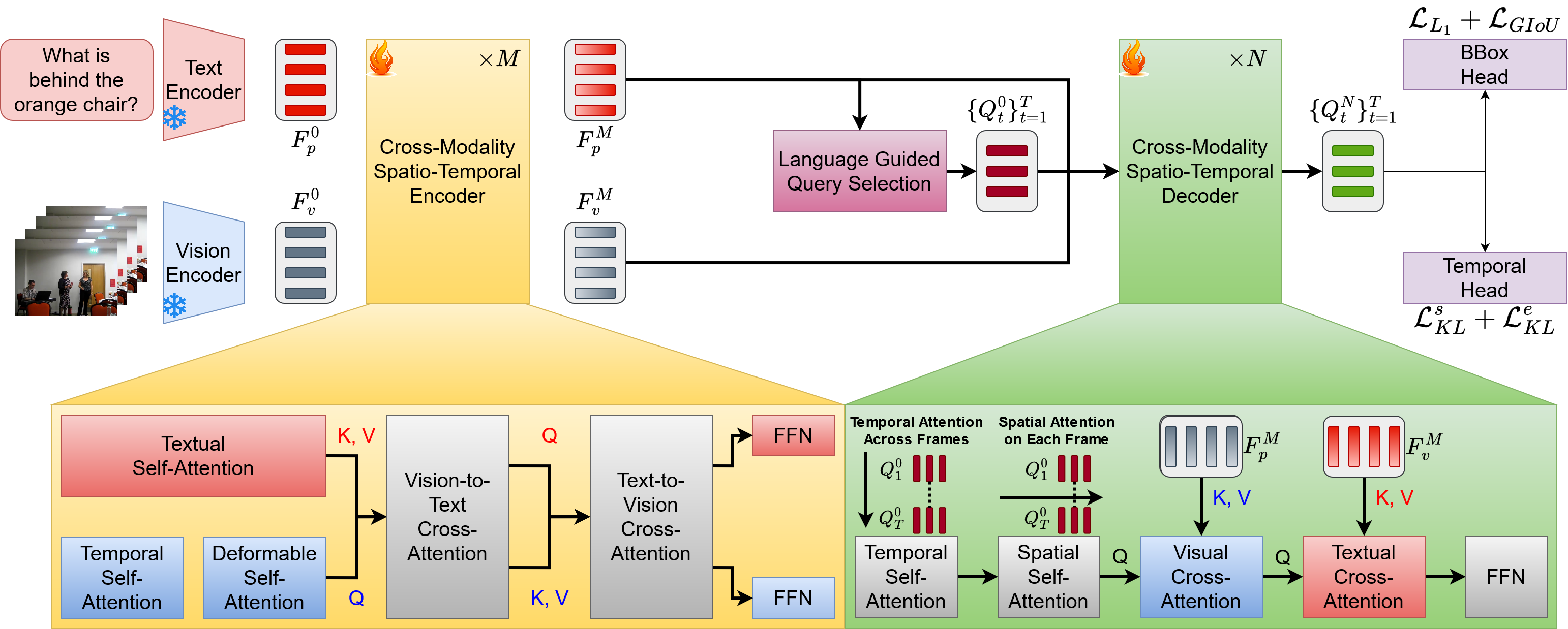}
    \caption{\textbf{Overall architecture:} We present our video grounding architecture. It consists of \textbf{vision and text encoders} that produce visual and textual features. A \textbf{cross-modality spatio-temporal encoder} which fuses information across spatial/temporal dimensions and visual/textual modalities. A \textbf{language guided query selction} module to initialize cross-modal queries. A \textbf{cross-modality spatio-temporal decoder} to decoder queries while fusing information from visual/textual features. And finally \textbf{two prediction heads} to predict the bounding boxes per frame and the temporal tube. Modules with (\protect\includegraphics[height=0.3cm]{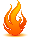}) are trainable and those with (\protect\includegraphics[height=0.3cm]{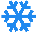}) are frozen.}
    \label{fig:main_arch}
\end{figure*}


\subsection{Problem Definition}
\label{sec:meth:problem}

The spatio-temporal video grounding task involves localizing and recognizing objects and actions in a video sequence by integrating spatial and temporal information.
In contrast to spatial grounding, which focuses on recognizing and localizing objects or actions within individual frames, spatio-temporal grounding extends this concept to include the temporal dimension. This means understanding where objects or actions are in each frame and how they evolve and move over time. 

Consider a video $V \in \R^{T \times H \times W \times C}$ with $T$ frames, $H\times W$ spatial resolution, and $C$ channels, respectively, along with a text prompt $P$.
The spatial grounding problem can be defined as the localization of one or more objects associated with the prompt $P$ in a frame $V_t, t \in \{1,...,T\}$ using a bounding box $B_i^t =(x_i^t, y_i^t, w_i^t, h_i^t)$, where $x_i^t$ and $y_i^t$ are the coordinates of the top-left corner, $w_i^t$ and $h_i^t$ are the width and height of the bounding box, $i \in \{1,...,N\}$ is the object number for the frame $t$. The temporal grounding problem, on the other hand, involves understanding how objects or actions evolve over time. It aims at localizing the temporal interval $(t_s, t_e)$ where the specific action/interaction happens in the entire temporal duration, where interval $(t_s, t_e)$ indicates the start and end frame of the object occurrence within the total frames $T$ ($1 \leq t_s < t_e \leq T$). Hence, the spatio-temporal grounding problem for object $i$ associated with prompt $P$ can be summarized as a set of spatio-temporal coordinates associated with the subset of frames where the object exists: $(x_i^t, y_i^t, w_i^t, h_i^t, t)$ and $ t \in \{t_s, ..., t_e\}$. The interval $(t_s, t_e) | \{1 \leq t_s < t_e \leq T\}$ and is a subset of the total frames $T$.

\subsection{Spatio-Temporal Video Grounding}
\label{sec:meth:video_grounding}

Here, we explain our video grounding model in \autoref{fig:main_arch}. As discussed earlier, we aim to design a spatio-temporal video grounding model that can perform well in closed-set and open-vocabulary settings. Strong open-vocabulary performance requires learning a rich visual/textual representation, which in turn requires a large amount of training data. Unfortunately, spatio-temporal video grounding datasets are quite limited in scale~\cite{zhang2020stgrnVIDSTG, tang2021stgvtHCSTVG}, resulting in current video grounding methods failing in generalizing well to out-of-distribution samples because they lack the requisite strong visual/textual representation.

To solve this problem, our approach takes inspiration from recent spatial grounding methods~\cite{huang2023kosmos-1, peng2023kosmos-2, chen2023shikra, you2023ferret, liu2023gdino}, which have strong open-vocabulary performance thanks to the large image-text corpus they are trained on. We can utilize the generalized representations of these models to enrich the weaker representation of video-grounding approaches obtained from the limited number of training samples. Our approach aims to leverage the strong pre-trained representations of spatial grounding methods to achieve strong closed-set supervised and open-vocabulary video grounding performance. Our spatio-temporal video grounding approach is based on the state-of-the-art DETR-based~\cite{carion2020detr} object detection framework DINO~\cite{zhang2022dino} and also borrows concepts of image-text alignment and grounding from Grounded Language-Image Pre-training (GLIP)~\cite{li2021glip} and Grounding DINO~\cite{liu2023gdino}. We extract initial features from backbone vision and text encoders $\theta_v$ and $\theta_p$. Following that, we model inter-frame and intra-frame features and learn cross-modal visual/textual relations in the Cross-Modality Spatio-Temporal Encoder (\autoref{sec:meth:video_grounding:encoder}). The result enriched cross-modal features are used to initialize queries for each frame (\autoref{sec:meth:video_grounding:query_select}). These queries are then decoded to predict the bounding boxes per frame and the temporal grounding start/end frame by aggregating information across spatial/temporal dimensions and injecting information through cross-attention from enriched visual/textual context (\autoref{sec:meth:video_grounding:decoder}).

Given the video $V$ and text prompt $P$ as described above, we obtain per-frame features $F^0_v$ and text features $F^0_p$ from vision and text encoders, $\theta_v$ and $\theta_p$, respectively. The vision encoder is based on a Swin Transformer~\cite{liu2021Swin}, and the text-encoder is defined as a BERT~\cite{devlin2019bert} model. Like other DETR-based detectors~\cite{zhu2021deformable, zhang2022dino}, image features are extracted from different vision-encoder blocks at multiple scales. The features are then passed to the Cross-Modality Spatio-Temporal Encoder.

\subsubsection{Cross-Modality Spatio-Temporal Encoder}
\label{sec:meth:video_grounding:encoder}

The initial vision and text features $F^0_v$ and $F^0_p$ neither contain any cross-modal information nor model the temporal relationship across frames. Therefore, we further encode the initial features through the Cross-Modality Spatio-Temporal Encoder to model the temporal information across frames and learn cross-modal features.

Each layer of the $M$ layer encoder first applies a Multi-Head Self-Attention ($\mathrm{MHSA}$)~\cite{vaswani2017attention} to the visual features $F_v$ along the temporal dimension, followed by a Deformable Attention ($\mathrm{DA}$)~\cite{zhu2021deformable} along the spatial dimension. This is done to model relations within frames and temporally across frames. Similarly, we apply a $\mathrm{MHSA}$ on the text features $F_p$. This is illustrated in \autoref{eq:meth:encoder:1}.
\begin{equation}
\begin{aligned}
    F^{m \prime}_v &= \mathrm{DA}^{m}_{spatial}(\mathrm{MHSA}^{m}_{temporal}(F^{m-1}_v)), \\
    F^{m \prime}_p &= \mathrm{MHSA}^{m}_p(F^{m-1}_p),
    \label{eq:meth:encoder:1}
\end{aligned}
\end{equation}
where $F^{m-1}_v$ and $F^{m-1}_p$ are visual and textual input features to layer $m$, $F^{m \prime}_v$ and $F^{m \prime}_p$ are the intermediate visual and textual feature representation, and $\mathrm{DA}^{m}_{spatial}$, $\mathrm{MHSA}^{m}_{temporal}$ and $\mathrm{MHSA}^{m}_p$ are the spatial-deformable, temporal and textual attentions at layer $m \in \{1,...,M\}$, respectively. Following initial spatial, temporal, and textual attentions, we fuse features across the visual and textual modalities, as done in GLIP~\cite{li2021glip}.

More specifically, we calculate the joint visual-textual attention, $\mathrm{Attn^m_{joint}}$, using projected intermediate features $F^{m \prime}_v$ and $F^{m \prime}_p$. This attention is then used alongside the intermediate features $F^{m \prime}_v$ and $F^{m \prime}_p$, to calculate the image-to-text and text-to-image cross-attentions as shown in \autoref{eq:meth:encoder:2} and \autoref{eq:meth:encoder:3}.
\begin{equation}
    \mathrm{Attn^m_{joint}} = \left( \frac{proj^m_{q,v}(F^{m \prime}_v)proj^m_{q,p}(F^{m \prime}_p)^T}{\sqrt{d^k}} \right),
    \label{eq:meth:encoder:2}
\end{equation}
where $proj^m_{q,v}$ and $proj^m_{q,p}$ are the query projections for the visual and textual features, respectively, at layer $m$.
\begin{equation}
\begin{aligned}
    F^{m}_v &= \mathrm{FFN}^{m}_v(\mathrm{softmax}(\mathrm{Attn^m_{joint}})proj^m_{p}(F^{m \prime}_p))), \\
    F^{m}_p &= \mathrm{FFN}^{m}_p(\mathrm{softmax}(\mathrm{Attn^m_{joint}}^T)proj^m_{v}(F^{m \prime}_v))),
    \label{eq:meth:encoder:3}
\end{aligned}
\end{equation}
where $F^{m}_v$ and $F^{m}_p$ are the final output features at layer $m$ of the encoder, $\mathrm{FFN}^{m}_v$ and $\mathrm{FFN}^{m}_v$ are the visual and textual Feed Forward Networks (FFN) and $proj^m_{v}$ and $proj^m_{p}$ are the linear layers to project the visual and textual features. The final encoded features at layer $m = M$ are then utilized to initialize cross-modal queries.

\subsubsection{Language-Guided Query Selection}
\label{sec:meth:video_grounding:query_select}

This module is designed to select features more relevant to the input text as decoder queries for effective language-vision fusion. We combine a DETR/DINO~\cite{carion2020detr, zhang2022dino} style queries with a sinusoidal temporal positional encoding to the positional part of the queries. The sinusoidal positional encoding added to the positional part of the queries adds important contextual information regarding the sequence of frames, allowing for improved temporal correlation and grounding~\cite{yang2022tubedetr}. The query selection module takes the encoder's visual and textual features as input and outputs $num\_query$ indices that correspond to the most relevant features for object detection per frame, $\{Q^0_t\}_{t=1}^T$, where $Q^0_t$ are the initial queries for the frame $t$. The module initializes the decoder queries using a combination of the selected indices and dynamic anchor boxes. The content part of the queries is set to be learnable during training, while the positional part is computed using the dynamic anchor boxes initialized using the encoder outputs. We also add a sinusoidal temporal positional encoding to the positional part of the queries.

\subsubsection{Cross-Modality Spatio-Temporal Decoder}
\label{sec:meth:video_grounding:decoder}

To decode the above queries into bounding box locations and temporal start/end tubes, we need to transform them into an output embedding, which can then be fed into prediction heads. The decoder allows for the queries to interact globally with others within a frame and across frames while utilizing the entire visual and textual features as context. Formally, the queries produced earlier are fed into a $N$ layer decoder. Each layer starts with a temporal self-attention, a spatial self-attention followed by a visual cross-attention and textual cross-attention, and finally an FFN. This is represented in \autoref{eq:meth:decoder:1}.
\begin{equation}
\begin{aligned}
    Q^{n \prime}_t &= \mathrm{MHSA}_{spatial}^{n}(\mathrm{MHSA}_{temporal}^{n}(Q^{n-1}_t)), \\
    Q^n_t &= \mathrm{FFN}^n(\mathrm{CA}^n_p(\mathrm{CA}^n_v(Q^{n \prime}_t, F^{M}_v), F^{M}_p)),
    \label{eq:meth:decoder:1}
\end{aligned}
\end{equation}
where $Q^{n-1}_t$ are the input queries at layer $n \in \{1,...,N\}$, $Q^{n \prime}_t$ are the intermediate queries after spatial and temporal attention at layer $n$, $Q^{n \prime}_t$ are the output queries at layer $n$, and $\mathrm{CA}^n_v$ and $\mathrm{CA}^n_p$ are the visual and textual cross-attentions at layer $n$. The cross-attentions are further elaborated in \autoref{eq:meth:decoder:2}.
\begin{equation}
\small
\begin{aligned}
    \mathrm{CA}^n_v(Q^{n \prime}_t, F^{M}_v) &= \left( \frac{proj^n_{q,v}(Q^{n \prime}_t)proj^n_{k,v}(F^{M}_v)^T}{\sqrt{d^k}}proj^n_{v}(F^{M}_v)^T \right), \\
    \mathrm{CA}^n_p(\mathrm{CA}^n_v, F^{M}_p) &= \left( \frac{proj^n_{q,p}(\mathrm{CA}^n_v)proj^n_{k,p}(F^{M}_p)^T}{\sqrt{d^k}}proj^n_{p}(F^{M}_p)^T \right),
    \label{eq:meth:decoder:2}
\end{aligned}
\end{equation}
where $proj^n_{q,v}$, $proj^n_{k,v}$ and $proj^n_{v}$ are the visual query, key and values projection for layer $n$ and $proj^n_{q,p}$, $proj^n_{k,p}$ and $proj^n_{p}$ are the textual query, key and value projections. The final queries from the decoder at layer $N$, $\{Q^N_t\}_{t=1}^T$, are then used for prediction.

\subsubsection{Prediction Heads}

The decoder outputs refined queries per frame $\{Q^N_t\}_{t=1}^T$. We follow the standard DETR-like bounding box regression head implemented as a Multi-Layer Perceptron (MLP), which predicts bounding boxes $B_i^t =(x_i^t, y_i^t, w_i^t, h_i^t)$, per frame. To predict the temporal interval $(t_s, t_e) | \{1 \leq t_s < t_e \leq T\}$, we add a temporal grounding head, implemented as an MLP, alongside the bounding box regression head, similar to existing works like~\cite{yang2022tubedetr, jin2022stcat}. The new head predicts the probabilities of the start $\tau_s \in [0,1]^T$ and ends $\tau_e \in [0,1]^T$ of the interval. During inference, the start and end interval $(t_s, t_e) | \{1 \leq t_s < t_e \leq T\}$ is computed by taking the maximum of the joint distribution of $(\tau_s, \tau_e)$. Any invalid combinations with $t_e \leq t_s$ are masked out.

\subsection{Loss Function}
\label{sec:meth:loss}

To leverage the generalized pre-trained representation from spatial-grounding foundation models, we initialize all spatial modules and cross attentions from the Grounding DINO~\cite{liu2023gdino} spatial grounding model. To preserve this generalized representation while ensuring effective modeling of the downstream task, we freeze the Vision and Text Encoders $\theta_v$ and $\theta_p$ and fine-tune the remaining components.

During training, the model receives a batch of videos $V$ with text prompt $P$. The ground-truth annotation contains the bounding box sequence $\{B_i^t\}_{t=t_s}^{t_e}$, and the corresponding start and end timestamps $(t_s, t_e)$. For spatial grounding, we follow the standard loss formulation used in DETR-like~\cite{carion2020detr, zhu2021deformable, zhang2022dino}, namely the $L_1$ loss, $\mathcal{L}_{L_1}$, and the Generalized Intersection over Union (GIoU)~\cite{rezatofighi2018giou} loss, $\mathcal{L}_{GIoU}$. Formally, the spatial grounding loss, $\mathcal{L}_{spatial}$ is defined in \autoref{eq:loss:spatial}.
\begin{equation}
    \mathcal{L}_{spatial} = \lambda_{L_1} \mathcal{L}_{L_1}(\hat{B},B) + \lambda_{GIoU} \mathcal{L}_{GIoU}(\hat{B},B).
    \label{eq:loss:spatial}
\end{equation}

For temporal grounding, we follow~\cite{rodriguez2020proposal,su2021stvgbert,yang2022tubedetr} and generate two $1$-dimensional gaussian heatmaps $\pi_s,\pi_e \in \mathcal{R}^T$, for the starting and ending positions. The temporal grounding loss is therefore defined in \autoref{eq:loss:temporal} as,
\begin{equation}
    \mathcal{L}_{temporal} = \mathcal{L}_{KL}^s(\hat{\pi_s},\pi_s) + \mathcal{L}_{KL}^e(\hat{\pi_e},\pi_e),
    \label{eq:loss:temporal}
\end{equation}
where $\mathcal{L}_{KL}^s$ and $\mathcal{L}_{KL}^e$ are the KL divergence losses for the start and end distributions, respectively.

Note that the model outputs the bounding boxes and starting/ending distributions during inference. We determine the temporal grounding segment, $(t_s, t_e)$, by taking the segment with the maximal joint start and end probability. Then, we consider the bounding boxes only within that tube for spatial grounding.

\section{Results}
\label{sec:results}

\begin{table*}[ht]
\caption{\small Performance comparisons of the state-of-the-art on HC-STVG V1~\cite{tang2021stgvtHCSTVG} and YouCook-Interactions~\cite{tan2021youcook-inter} in open-vocabulary setting.}
\centering
        \begin{tabular}{lccccc}
        \toprule
        \multirow{2}{*}{Method} & \multirow{2}{*}{Pre-training} & \multicolumn{3}{c}{HC-STVG V1} & YouCook-Interactions \\
        \cmidrule(l){3-6} 
                                                             &           & m\_vIoU  & vIoU@0.3 & vIoU@0.5 & Accuracy \\
        \midrule
        TubeDETR~\textit{(CVPR'22)}~\cite{yang2022tubedetr}  & VidSTG    & 16.84    & 22.32    & 9.22    & 51.63  \\
        STCAT~\textit{(NeurIPS'22)}~\cite{jin2022stcat}      & VidSTG    & 22.58    & 32.14    & 20.83    & 55.90  \\
        \rowcolor{violet!10} 
        VideoGrounding-DINO                                  & VidSTG    & \textbf{27.46} & \textbf{40.13} & \textbf{29.92} & \textbf{57.73} \\
        \bottomrule
    \end{tabular}
\label{tab:open}
\vspace{-0.3cm}
\end{table*}

\subsection{Experimental Setup and Protocols}
\label{sec:results:setup}

Below, we first briefly explain the implementation details (\autoref{sec:results:setup:implementation}), followed by evaluation settings (\autoref{sec:results:setup:evaluation}), and datasets (\autoref{sec:results:setup:datasets}) used in our work.

\subsubsection{Implementation Details}
\label{sec:results:setup:implementation}

As discussed in the methodology (\autoref{sec:method}), we initialize the spatial modules in our model from the Grounding DINO~\cite{liu2023gdino} spatial grounding model and keep the vision and text encoders frozen. Our prediction heads for both spatial and temporal predictions are set to be 3-layer Multi-Layer Perceptrons (MLPs). We sample 128 frames during training and inference, resized to a resolution of $448$ on the shorter side. We set both $M$ and $N$ to $6$, and train the model with a batch size of $8$ and learning rate of $1e^{-4}$, and weight decay if $10^{-4}$. The number of epochs for VidSTG is set to $10$, and for HC-STVG V1/V2 is set to $90$.

\subsubsection{Evaluation Settings}
\label{sec:results:setup:evaluation}

We evaluate our video grounding model in two settings, \emph{Open-Vocabulary} and \emph{Closed-Set Supervised}.

\noindent \textbf{Open-Vocabulary Evaluation:} In the open-vocabulary setting we train our model on the VidSTG~\cite{zhang2020stgrnVIDSTG} dataset and then evaluate on two different datasets, HC-STVG V1~\cite{tang2021stgvtHCSTVG} and YouCook-Interactions~\cite{tan2021youcook-inter} to understand how well the model generalizes to new distributions. The reason for choosing these two datasets is that the former provides a relatively minor distribution shift given the similar perspective/objects in the videos compared to the training dataset VidSTG. In contrast, the latter provides a major distribution shift with changes in perspective and annotated objects/interactions.

\noindent \textbf{Closed-Set Supervised Evaluation:} In the supervised evaluation setting, we train on the training set and evaluate each dataset's respective validation/testing set. This evaluation is conducted for three majorly used datasets in spatio-temporal video grounding, namely VidSTG~\cite{zhang2020stgrnVIDSTG}, HC-STVG V1~\cite{tang2021stgvtHCSTVG} and HC-STVG V2~\cite{tang2021stgvtHCSTVG}.

\begin{table*}[ht]
\caption{\small Performance comparisons of the state-of-the-art on the VidSTG~\cite{zhang2020stgrnVIDSTG} test set in closed-set supervised setting.}
\centering\setlength{\tabcolsep}{4pt}
\resizebox{\textwidth}{!}{
    \begin{tabular}{lcccccccc}
    \toprule
    \multirow{2}{*}{Method} & \multicolumn{4}{c}{Declarative Sentences} & \multicolumn{4}{c}{Interrogative Sentences} \\
    \cmidrule(l){2-9} 
                        & m\_tIoU  & m\_vIoU  & vIoU@0.3 & vIoU@0.5 & \multicolumn{1}{l}{m\_tIoU} & \multicolumn{1}{c}{m\_vIoU} & \multicolumn{1}{c}{vIoU@0.3} & \multicolumn{1}{c}{vIoU@0.5} \\
        \midrule
        \textbf{\textit{Factorized:}}  &  &   &  &  &   &  &   &    \\
        GroundeR~\textit{(ECCV'16)}~\cite{rohrbach2016grounding}+TALL~\textit{(ICCV'17)}~\cite{gao2017tall} & \multirow{3}{*}{34.63}  & 9.78  & 11.04 & 4.09 & \multirow{3}{*}{33.73}  & 9.32 &  11.39 & 3.24   \\
        STPR~\textit{(ICCV'17)}~\cite{yamaguchi2017spatio}+TALL~\textit{(ICCV'17)}~\cite{gao2017tall} &  &  10.40 & 12.38 & 4.27 &   & 9.98 & 11.74  & 4.36   \\
        WSSTG~\textit{(arXiv'19)}~\cite{chen2019weakly}+TALL~\textit{(ICCV'17)}~\cite{gao2017tall} &  &  11.36 & 14.63 & 5.91 &   & 10.65 & 13.90  & 5.32   \\
        GroundeR~\textit{(ECCV'16)}~\cite{rohrbach2016grounding}+L-Net~\textit{(AAAI'19)}~\cite{chen2019localizing} & \multirow{3}{*}{40.86}  &  11.89 & 15.32 & 5.45 &  \multirow{3}{*}{39.79} & 11.05 & 14.28  & 5.11   \\
        STPR~\textit{(ICCV'17)}~\cite{yamaguchi2017spatio}+L-Net~\textit{(AAAI'19)}~\cite{chen2019localizing} &  &  12.93 & 16.27 & 5.68 &  & 11.94  & 14.73 & 5.27      \\
        WSSTG~\textit{(arXiv'19)}~\cite{chen2019weakly}+L-Net~\textit{(AAAI'19)}~\cite{chen2019localizing} &  &  14.45 & 18.00 & 7.89 &   & 13.36  & 17.39  & 7.06   \\
        \midrule
        \textbf{\textit{Two-Stage:}}  &  &   &  &  &   &  &   &    \\
        STGRN~\textit{(CVPR'20)}~\cite{zhang2020stgrnVIDSTG} &  48.47 &  19.75 & 25.77 & 14.60 &  46.98 & 18.32 & 21.10  & 12.83   \\
        STGVT~\textit{(TCSVT'21)}~\cite{tang2021stgvtHCSTVG} &  - &  21.62 & 29.80 & 18.94 &  - & - & -  & -   \\
        OMRN~\textit{(IJCAI'21)}~\cite{zhang2022omrn} &  50.73 &  23.11 & 32.61 & 16.42 &  49.19 & 20.63 & 28.35  & 14.11   \\
        \midrule
        \textbf{\textit{One-Stage:}}  &  &   &  &  &   &  &   &    \\
        STVGBert~\textit{(ICCV'21)}~\cite{su2021stvgbert} & - &  23.97 & 30.91 & 18.39 & - & 22.51 & 25.97  & 15.95   \\
        TubeDETR~\textit{(CVPR'22)}~\cite{yang2022tubedetr} & 48.10 &  30.40 & 42.50 & 28.20 & 46.90 & 25.70 & 35.70  & 23.20   \\
        STCAT~\textit{(NeurIPS'22)}~\cite{jin2022stcat} &  50.82 & 33.14  & 46.20 & 32.58 & 49.67 & 28.22 & 39.24 & 26.63 \\
        STVGFormer~\textit{(CVPR'23)}~\cite{lin2023stvgformer} &  - & 33.70  & 47.20 & 32.80 & - & 28.50 & 39.90 & 26.20 \\
        \rowcolor{violet!10} 
        VideoGrounding-DINO &  \textbf{51.97} & \textbf{34.67}  & \textbf{48.11} & \textbf{33.96} & \textbf{50.83} & \textbf{29.89} & \textbf{41.03} & \textbf{27.58} \\
        \bottomrule
    \end{tabular}}
\label{tab:sup:vidstg}
\vspace{-0.3cm}
\end{table*}

\subsubsection{Datasets}
\label{sec:results:setup:datasets}

We evaluate our approach and compare against the state-of-the-art in two settings: \emph{Open-Vocabulary} and \emph{Closed-Set Supervised}, across a total of four grounding datasets, namely: VidSTG~\cite{zhang2020stgrnVIDSTG}, HCSTVG V1~\cite{tang2021stgvtHCSTVG}, HCSTVG V2~\cite{tang2021stgvtHCSTVG}, and YouCook-Interactions~\cite{tan2021youcook-inter}.

\begin{table}[ht]
\caption{\small Performance comparisons of the state-of-the-art on the HC-STVG V1~\cite{tang2021stgvtHCSTVG} test set in closed-set supervised setting.}
\centering\setlength{\tabcolsep}{4pt}
    \resizebox{\columnwidth}{!}{
    \begin{tabular}{lccc}
        \toprule
        Methods & m\_vIoU & vIoU@0.3 &  vIoU@0.5  \\
        \midrule
        STGVT~\textit{(TCSVT'21)}~\cite{tang2021stgvtHCSTVG} &  18.15 & 26.81 & 9.48  \\
        STVGBert~\textit{(ICCV'21)}~\cite{su2021stvgbert} & 20.42  & 29.37 &  11.31  \\
        TubeDETR~\textit{(CVPR'22)}~\cite{yang2022tubedetr} & 32.40  & 49.80 & 23.50   \\
        STCAT~\textit{(NeurIPS'22)}~\cite{jin2022stcat}  & 35.09  & 57.67 & 30.09 \\
        STVGFormer~\textit{(CVPR'23)}~\cite{lin2023stvgformer}  & 36.90  & 62.20 & 34.80 \\
        \rowcolor{violet!10} 
        VideoGrounding-DINO & \textbf{38.25}  & \textbf{62.47} & \textbf{36.14} \\
        \bottomrule
    \end{tabular}}
\label{tab:sup:hcstvg_v1}
\vspace{-0.2 cm}
\end{table}
\begin{table}[ht]
\caption{\small Performance comparisons of the state-of-the-art on the HC-STVG V2~\cite{tang2021stgvtHCSTVG} val set in closed-set supervised setting.}
\centering\setlength{\tabcolsep}{4pt}
    \resizebox{\columnwidth}{!}{
    \begin{tabular}{lccc}
        \toprule
        Methods & m\_vIoU & vIoU@0.3 &  vIoU@0.5  \\
        \midrule
        Yu \textit{et al}~\textit{(arXiv'21)}~\cite{yu2022yu2rd} & 30.00  & - &  -  \\
        Aug. 2D-TAN~\textit{(arXiv'21)}~\cite{tan2021aug2d}  & 30.40  & 50.40 & 18.80 \\
        TubeDETR~\textit{(CVPR'22)}~\cite{yang2022tubedetr} & 36.40  & 58.80 & 30.60 \\
        STVGFormer~\textit{(CVPR'23)}~\cite{lin2023stvgformer}  & 38.70  & 65.50 & 33.80 \\
       \rowcolor{violet!10}  VideoGrounding-DINO & \textbf{39.88}  & \textbf{67.13} & \textbf{34.49} \\
        \bottomrule
    \end{tabular}}
\label{tab:sup:hcstvg_v2}
\vspace{-0.5 cm}
\end{table}

\noindent \textbf{VidSTG:} The VidSTG~\cite{zhang2020stgrnVIDSTG} dataset is derived from the VidOR~\cite{shang2019vidor} dataset, incorporating object relation annotations. It includes 99,943 video-text pairs, encompassing 44,808 declarative sentence queries and 55,135 interrogative sentence queries. The training, validation, and test sets consist of 80,684, 8,956, and 10,303 sentences and 5,436, 602, and 732 videos, respectively. VidSTG's text queries are confined to describing pre-defined object/relation categories in VidOR~\cite{shang2019vidor}.

\noindent \textbf{HC-STVG V1/V2:} The HC-STVG datasets are sourced from movie scenes, each video clip spanning approximately 20 seconds. These datasets pose challenges in spatio-temporal grounding due to video clips featuring multiple individuals engaged in similar actions. HC-STVG V1 comprises 4,500 training and 1,160 testing video-text pairs. HC-STVG V2 expands HC-STVG V1, enhancing annotation quality with 10,131, 2,000, and 4,413 samples for training, validation, and testing, respectively. As HC-STVG V2's test set annotations are unavailable publicly, results are reported on the validation set.

\begin{table*}[ht]
\caption{\small Ablation on various design choices for our approach on the VidSTG~\cite{zhang2020stgrnVIDSTG} test set in closed-set supervised setting.}
\centering
\resizebox{\textwidth}{!}{
    \begin{tabular}{@{}lcccccccc@{}}
    \toprule
    \multirow{2}{*}{Method} & \multicolumn{4}{c}{Declarative Sentences} & \multicolumn{4}{c}{Interrogative Sentences} \\
    \cmidrule(l){2-9} 
                        & m\_tIoU  & m\_vIoU  & vIoU@0.3 & vIoU@0.5 & \multicolumn{1}{l}{m\_tIoU} & \multicolumn{1}{c}{m\_vIoU} & \multicolumn{1}{c}{vIoU@0.3} & \multicolumn{1}{c}{vIoU@0.5} \\
        \midrule
        Naive Solution (Frozen Grounding DINO~\cite{liu2023gdino}) & 39.78 & 18.07 & 22.31 & 13.75 & 39.79 & 9.66  & 10.42 & 3.84  \\
        + Decoder Temporal Aggregation                    & 42.81 & 20.74 & 26.53 & 15.41 & 43.81 & 12.38 & 16.71 & 8.62  \\
        + Encoder Temporal Aggregation                    & 46.29 & 23.19 & 32.38 & 18.95 & 47.17 & 16.19 & 23.28 & 13.04 \\
        + Finetuned Spatial Modules in Decoder            & 48.06 & 28.97 & 41.60 & 26.06 & 49.58 & 24.27 & 32.85 & 20.11 \\
        + Finetuned Spatial Modules in Encoder            & 51.97 & 34.67 & 48.11 & 33.96 & 50.83 & 29.89 & 41.03 & 27.58 \\
        \bottomrule
    \end{tabular}}
\label{tab:abl}
\end{table*}

\noindent \textbf{YouCook-Interactions:} The YouCook-Interactions~\cite{tan2021youcook-inter} dataset serves as an expansion of the YouCook2~\cite{zhou2018youcook2} dataset focused on cooking instructions. This extension includes bounding boxes for 6,000 carefully chosen frames, typically encompassing the hand and the tool specified in the corresponding sentence-level annotations. Our assessment revolves around examining models' spatial grounding capabilities using this dataset.

\subsection{Experimental Results and Analysis}
\label{sec:results:analysis}

In this section, we present our results across the evaluation mentioned above settings (\autoref{sec:results:setup:evaluation}) and datasets (\autoref{sec:results:setup:datasets}). We start with the \emph{closed-set} evaluation in \autoref{sec:results:analysis:supervised}, followed by the \emph{open-vocabulary} evaluation in \autoref{sec:results:analysis:open}.

\subsubsection{Open-Vocabulary Evaluation}
\label{sec:results:analysis:open}

For open-vocabulary evaluation, we train on VidSTG~\cite{zhang2020stgrnVIDSTG} and present results HC-STVG V1~\cite{tang2021stgvtHCSTVG} and YouCook-Interactions~\cite{tan2021youcook-inter}. The results are reported jointly in \autoref{tab:open}.

\noindent \textbf{Results on HC-STVG V1:} We report open-vocabulary evaluation on m\_vIoU, vIoU@0.3, and vIoU@0.5. We achieve state-of-the-art performance over both TubeDETR~\cite{yang2022tubedetr} and STCAT~\cite{jin2022stcat}. We attribute this strong performance to our design, which leverages the strong pre-trained generalized features of a spatial grounding foundation model.

\noindent \textbf{Results on YouCook-Interactions:} We evaluate our method further on the YouCook-Interactions~\cite{tan2021youcook-inter} dataset, reporting pointing game accuracy for spatial grounding. Our approach gains nearly $2\%$ in accuracy over STCAT~\cite{jin2022stcat} and more than $6\%$ compared to TubeDETR~\cite{yang2022tubedetr}. This further shows our strong generalization capabilities in the open-vocabulary setting.

\begin{figure*}[ht]
\centering
    \includegraphics[width=\textwidth]{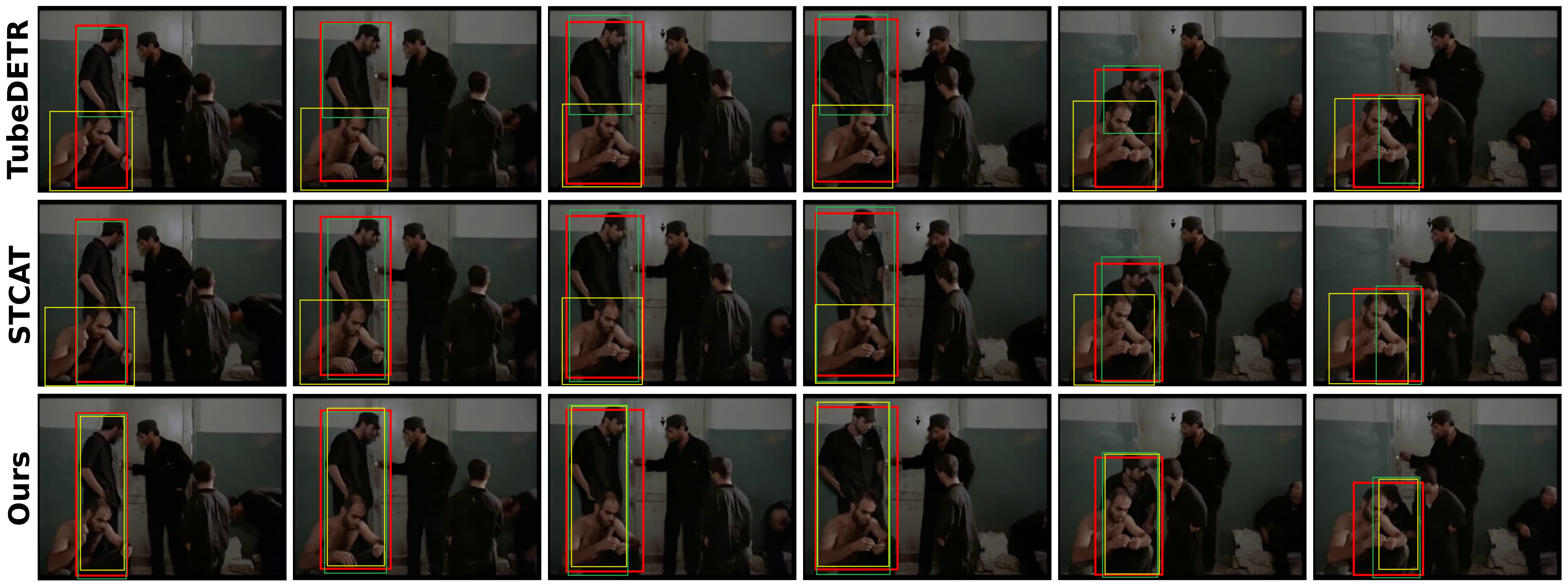}
    \vspace{-.6cm}
    \caption{Sample visualization for video grounding result on HC-STVG V1~\cite{tang2021stgvtHCSTVG} for TubeDETR~\cite{yang2022tubedetr}, STCAT~\cite{jin2022stcat} and ours with the prompt \textbf{The man behind the shirtless man turns and squats}. We show bounding boxes for \textcolor{red}{Ground-truth}, \textcolor{green}{Closed-Set Supervised}, and \textcolor{myYellow}{Open-Vocabulary} results. Note how both TubeDETR and STCAT are close to the ground truth in the supervised setting (STCAT more so than TubeDETR), they cannot correctly ground the text properly in the open-vocabulary setting.}
    \label{fig:vis}
    \vspace{-0.3cm}
\end{figure*}

\subsubsection{Closed-Set Supervised Evaluation}
\label{sec:results:analysis:supervised}

We present closed-set evaluations across three datasets, VidSTG~\cite{zhang2020stgrnVIDSTG}, HC-STVG V1~\cite{tang2021stgvtHCSTVG} and HC-STVG V2~\cite{tang2021stgvtHCSTVG}.

\noindent \textbf{Results on VidSTG:} We present results on the VidSTG test set in closed-set setting in \autoref{tab:sup:vidstg}, reporting m\_tIoU, m\_vIoU, vIoU@0.3 and vIoU@0.5. The results show that our method achieves state-of-the-art performance in comparison to both \emph{Two-Stage} and \emph{One-Stage} methods. In particular, we achieve more than $1 t\_IoU$ gain in temporal grounding over the previous best methods OMRN~\cite{zhang2022omrn} (\emph{One-Stage}) and STVGFormer~\cite{lin2023stvgformer} (\emph{Two-Stage}), both for Declarative and Interrogative sentences. Similarly, for m\_vIoU, vIoU@0.3 and vIoU@0.5, we achieve a more than $1$ unit gain the state-of-the-art methods STVGFormer~\cite{lin2023stvgformer} and STCAT~\cite{jin2022stcat}. Note that our method uses a frozen visual and textual encoder. In contrast, those mentioned above previous state-of-the-art methods all train the entire encoder.

\noindent \textbf{Results on HC-STVG V1:} We present results on the HC-STVG V1 dataset in \autoref{tab:sup:hcstvg_v1}, reporting m\_vIoU, vIoU@0.3 and vIoU@0.5. We achieve a nearly $1.5$ unit gain in m\_vIoU and vIoU@0.5 and a $1$ unit gain in vIoU@0.3 over the previous best method STVGFormer~\cite{lin2023stvgformer}. This shows the consistent performance of our method on this dataset.

\noindent \textbf{Results on HC-STVG V2:} We present results on the HC-STVG V2 dataset in \autoref{tab:sup:hcstvg_v2}, reporting m\_vIoU, vIoU@0.3 and vIoU@0.5. Our performance gain on HC-STVG V1 is reflected here as well, with a consistent performance gain on HC-STVG V2, in comparison to the SoTA~\cite{lin2023stvgformer,yang2022tubedetr}.

\subsection{Ablative Analysis}

We perform an ablative analysis of the various design choices for our model. In particular, we first evaluate a naive baseline where no additional temporal aggregators are added, and all pre-trained spatial modules are frozen in the encoder and decoder. This baseline is relatively weak in both temporal and spatial grounding. We next add temporal modules in first the decoder, and subsequently in the encoder. We find that it gives significant improvements in the temporal grounding, and additionally also improves the spatial grounding. Finally, we finetune the pre-trained spatial modules in both the decoder and the encoder, which provides a strong improvement in spatial grounding, alongside consistent improvement in temporal grounding.

\subsection{Limitation}
While our video grounding model excels in closed-set and open-vocabulary scenarios, it leverages image-text pre-trained models like Grounding DINO~\cite{liu2023gdino}. To enhance understanding in open-vocabulary settings, an extension to video-language pre-training on a larger and more diverse dataset, akin to CLIP~\cite{radford2021clip}, can help further boost performance. Building a video-language pre-training dataset with diverse natural language expressions and spatio-temporal localization is imperative, given the constraints of datasets like VidSTG~\cite{zhang2020stgrnVIDSTG} and HC-STVG~\cite{tang2021stgvtHCSTVG}.
\section{Conclusion}
\label{sec:conclusion}

This paper introduces an Open-Vocabulary Spatio-Temporal Video Grounding task, enhancing current closed-set methodologies by using pre-trained representations from spatial grounding models. The proposed model performs well in closed-set and open-vocabulary scenarios, surpassing state-of-the-art results in supervised setting on VidSTG and HC-STVG datasets, and outperforming recent models in open-vocabulary on HC-STVG V1 and YouCook-Interactions. Its architecture includes learnable adapter blocks for video-specific adaptation, bridging the semantic gap between natural language queries and visual content. This research addresses open-vocabulary challenges and explores achieving robust performance without extensive video annotations, paving the way for open-vocabulary video grounding.

{
    \small
    \bibliographystyle{ieeenat_fullname}
    \bibliography{main}

\begin{thebibliography}{35}
\providecommand{\natexlab}[1]{#1}
\providecommand{\url}[1]{\texttt{#1}}
\expandafter\ifx\csname urlstyle\endcsname\relax
  \providecommand{\doi}[1]{doi: #1}\else
  \providecommand{\doi}{doi: \begingroup \urlstyle{rm}\Url}\fi

\bibitem[Carion et~al.(2020)Carion, Massa, Synnaeve, Usunier, Kirillov, and Zagoruyko]{carion2020detr}
Nicolas Carion, Francisco Massa, Gabriel Synnaeve, Nicolas Usunier, Alexander Kirillov, and Sergey Zagoruyko.
\newblock End-to-end object detection with transformers.
\newblock In \emph{ECCV}, 2020.

\bibitem[Chen et~al.(2023{\natexlab{a}})Chen, Shvetsova, Rouditchenko, Kondermann, Thomas, Chang, Feris, Glass, and Kuehne]{chen2023what}
Brian Chen, Nina Shvetsova, Andrew Rouditchenko, Daniel Kondermann, Samuel Thomas, Shih-Fu Chang, Rogerio Feris, James Glass, and Hilde Kuehne.
\newblock What, when, and where? -- self-supervised spatio-temporal grounding in untrimmed multi-action videos from narrated instructions.
\newblock \emph{arXiv preprint arXiv:2303.16990}, 2023{\natexlab{a}}.

\bibitem[Chen et~al.(2019{\natexlab{a}})Chen, Ma, Chen, Jie, , and Luo]{chen2019localizing}
Jingyuan Chen, Lin Ma, Xinpeng Chen, Zequn Jie, , and Jiebo Luo.
\newblock Localizing natural language in videos.
\newblock In \emph{AAAI}, 2019{\natexlab{a}}.

\bibitem[Chen et~al.(2023{\natexlab{b}})Chen, Zhang, Zeng, Zhang, Zhu, and Zhao]{chen2023shikra}
Keqin Chen, Zhao Zhang, Weili Zeng, Richong Zhang, Feng Zhu, and Rui Zhao.
\newblock Shikra: Unleashing multimodal llm's referential dialogue magic.
\newblock \emph{arXiv preprint arXiv:2306.15195}, 2023{\natexlab{b}}.

\bibitem[Chen et~al.(2019{\natexlab{b}})Chen, Ma, Luo, , and Wong]{chen2019weakly}
Zhenfang Chen, Lin Ma, Wenhan Luo, , and Kwan-Yee~K Wong.
\newblock Weakly-supervised spatio-temporally grounding natural sentence in video.
\newblock \emph{arXiv preprint arXiv:1906.02549}, 2019{\natexlab{b}}.

\bibitem[Devlin et~al.(2019)Devlin, Chang, Lee, and Toutanova]{devlin2019bert}
Jacob Devlin, Ming-Wei Chang, Kenton Lee, and Kristina Toutanova.
\newblock Bert: Pre-training of deep bidirectional transformers for language understanding.
\newblock \emph{arXiv preprint arXiv:1810.04805}, 2019.

\bibitem[Gao et~al.(2017)Gao, Sun, Yang, , and Nevatia]{gao2017tall}
Jiyang Gao, Chen Sun, Zhenheng Yang, , and Ram Nevatia.
\newblock Tall: Temporal activity localization via language query.
\newblock In \emph{ICCV}, 2017.

\bibitem[Huang et~al.(2023)Huang, Dong, Wang, Hao, Singhal, Ma, Lv, Cui, Mohammed, Liu, Aggarwal, Chi, Bjorck, Chaudhary, Som, Song, and Wei]{huang2023kosmos-1}
Shaohan Huang, Li Dong, Wenhui Wang, Yaru Hao, Saksham Singhal, Shuming Ma, Tengchao Lv, Lei Cui, Owais~Khan Mohammed, Qiang Liu, Kriti Aggarwal, Zewen Chi, Johan Bjorck, Vishrav Chaudhary, Subhojit Som, Xia Song, and Furu Wei.
\newblock Language is not all you need: Aligning perception with language models.
\newblock \emph{arXiv preprint arXiv:2302.14045}, 2023.

\bibitem[Jin et~al.(2022)Jin, Li, Yuan, and Mu]{jin2022stcat}
Yang Jin, Yongzhi Li, Zehuan Yuan, and Yadong Mu.
\newblock Embracing consistency: A one-stage approach for spatio-temporal video grounding.
\newblock In \emph{NeurIPS}, 2022.

\bibitem[Li* et~al.(2022)Li*, Zhang*, Zhang*, Yang, Li, Zhong, Wang, Yuan, Zhang, Hwang, Chang, and Gao]{li2021glip}
Liunian~Harold Li*, Pengchuan Zhang*, Haotian Zhang*, Jianwei Yang, Chunyuan Li, Yiwu Zhong, Lijuan Wang, Lu Yuan, Lei Zhang, Jenq-Neng Hwang, Kai-Wei Chang, and Jianfeng Gao.
\newblock Grounded language-image pre-training.
\newblock In \emph{CVPR}, 2022.

\bibitem[Lin et~al.(2023)Lin, Tan, Hu, Jin, Ye, and Zheng]{lin2023stvgformer}
Zihang Lin, Chaolei Tan, Jian-Fang Hu, Zhi Jin, Tiancai Ye, and Wei-Shi Zheng.
\newblock Collaborative static and dynamic vision-language streams for spatio-temporal video grounding.
\newblock In \emph{CVPR}, 2023.

\bibitem[Liu et~al.(2023)Liu, Zeng, Ren, Li, Zhang, Yang, Li, Yang, Su, Zhu, et~al.]{liu2023gdino}
Shilong Liu, Zhaoyang Zeng, Tianhe Ren, Feng Li, Hao Zhang, Jie Yang, Chunyuan Li, Jianwei Yang, Hang Su, Jun Zhu, et~al.
\newblock Grounding dino: Marrying dino with grounded pre-training for open-set object detection.
\newblock \emph{arXiv preprint arXiv:2303.05499}, 2023.

\bibitem[Liu et~al.(2021)Liu, Lin, Cao, Hu, Wei, Zhang, Lin, and Guo]{liu2021Swin}
Ze Liu, Yutong Lin, Yue Cao, Han Hu, Yixuan Wei, Zheng Zhang, Stephen Lin, and Baining Guo.
\newblock Swin transformer: Hierarchical vision transformer using shifted windows.
\newblock In \emph{ICCV}, 2021.

\bibitem[Peng et~al.(2023)Peng, Wang, Dong, Hao, Huang, Ma, and Wei]{peng2023kosmos-2}
Zhiliang Peng, Wenhui Wang, Li Dong, Yaru Hao, Shaohan Huang, Shuming Ma, and Furu Wei.
\newblock Kosmos-2: Grounding multimodal large language models to the world.
\newblock \emph{arXiv preprint arXiv:2306.14824}, 2023.

\bibitem[Radford et~al.(2021)Radford, Kim, Hallacy, Ramesh, Goh, Agarwal, Sastry, Askell, Mishkin, Clark, Krueger, and Sutskever]{radford2021clip}
Alec Radford, Jong~Wook Kim, Chris Hallacy, Aditya Ramesh, Gabriel Goh, Sandhini Agarwal, Girish Sastry, Amanda Askell, Pamela Mishkin, Jack Clark, Gretchen Krueger, and Ilya Sutskever.
\newblock Learning transferable visual models from natural language supervision.
\newblock In \emph{ICML}, 2021.

\bibitem[Rasheeda et~al.(2023)Rasheeda, Maaz, Shaji, Shaker, Khan, Cholakkal, Anwer, Xing, Yang, and Khan]{hanoona2023GLaMM}
Hanoona Rasheeda, Muhammad Maaz, Sahal Shaji, Abdelrahman Shaker, Salman Khan, Hisham Cholakkal, Rao~M. Anwer, Eric Xing, Ming-Hsuan Yang, and Fahad~S. Khan.
\newblock Glamm: Pixel grounding large multimodal model.
\newblock \emph{arXiv preprint arXiv:2311.03356}, 2023.

\bibitem[Rezatofighi et~al.(2019)Rezatofighi, Tsoi, Gwak, Sadeghian, Reid, and Savarese]{rezatofighi2018giou}
Hamid Rezatofighi, Nathan Tsoi, JunYoung Gwak, Amir Sadeghian, Ian Reid, and Silvio Savarese.
\newblock Generalized intersection over union.
\newblock In \emph{CVPR}, 2019.

\bibitem[Rodriguez et~al.(2020)Rodriguez, Marrese-Taylor, Saleh, Li, and Gould]{rodriguez2020proposal}
Cristian Rodriguez, Edison Marrese-Taylor, Fatemeh~Sadat Saleh, Hongdong Li, and Stephen Gould.
\newblock Proposal-free temporal moment localization of a natural-language query in video using guided attention.
\newblock In \emph{WACV}, 2020.

\bibitem[Rohrbach et~al.(2016)Rohrbach, Rohrbach, Hu, Darrell, , and Schiele]{rohrbach2016grounding}
Anna Rohrbach, Marcus Rohrbach, Ronghang Hu, Trevor Darrell, , and Bernt Schiele.
\newblock Grounding of textual phrases in images by reconstruction.
\newblock In \emph{ECCV}, 2016.

\bibitem[Shang et~al.(2019)Shang, Di, Xiao, Cao, Yang, and Chua]{shang2019vidor}
Xindi Shang, Donglin Di, Junbin Xiao, Yu Cao, Xun Yang, and Tat-Seng Chua.
\newblock Annotating objects and relations in user-generated videos.
\newblock In \emph{ICMR}, 2019.

\bibitem[Su et~al.(2021)Su, Yu, , and Xu]{su2021stvgbert}
Rui Su, Qian Yu, , and Dong Xu.
\newblock Stvgbert: A visual- linguistic transformer based framework for spatio-temporal video grounding.
\newblock In \emph{ICCV}, 2021.

\bibitem[Tan et~al.(2021{\natexlab{a}})Tan, Lin, Hu, Li, and Zheng]{tan2021aug2d}
Chaolei Tan, Zihang Lin, Jian-Fang Hu, Xiang Li, and Wei-Shi Zheng.
\newblock Augmented 2d-tan: A two-stage approach for human-centric spatio-temporal video grounding.
\newblock \emph{arXiv preprint arXiv:2106.10634}, 2021{\natexlab{a}}.

\bibitem[Tan et~al.(2021{\natexlab{b}})Tan, Plummer, Saenko, Jin, and Russell]{tan2021youcook-inter}
Reuben Tan, Bryan~A. Plummer, Kate Saenko, Hailin Jin, and Bryan Russell.
\newblock Look at what i'm doing: Self-supervised spatial grounding of narrations in instructional videos, 2021{\natexlab{b}}.

\bibitem[Tang et~al.(2021)Tang, Liao, Liu, Li, Jin, Jiang, Yu, and Xu]{tang2021stgvtHCSTVG}
Zongheng Tang, Yue Liao, Si Liu, Guanbin Li, Xiaojie Jin, Hongxu Jiang, Qian Yu, and Dong Xu.
\newblock Human-centric spatio-temporal video grounding with visual transformers.
\newblock \emph{IEEE Transactions on Circuits and Systems for Video Technology}, 2021.

\bibitem[Vaswani et~al.(2017)Vaswani, Shazeer, Parmar, Uszkoreit, Jones, Gomez, Kaiser, and Polosukhin]{vaswani2017attention}
Ashish Vaswani, Noam Shazeer, Niki Parmar, Jakob Uszkoreit, Llion Jones, Aidan~N Gomez, \L~ukasz Kaiser, and Illia Polosukhin.
\newblock Attention is all you need.
\newblock In \emph{NeurIPS}, 2017.

\bibitem[Wang et~al.(2022)Wang, Wu, Li, and Wu]{wang2022mmn}
Zhenzhi Wang, Tao Wu, Tianhao Li, and Gangshan Wu.
\newblock Negative sample matters: {A} renaissance of metric learning for temporal grounding.
\newblock In \emph{AAAI}, 2022.

\bibitem[Yamaguchi et~al.(2017)Yamaguchi, Saito, Ushiku, , and Harada]{yamaguchi2017spatio}
Masataka Yamaguchi, Kuniaki Saito, Yoshitaka Ushiku, , and Tatsuya Harada.
\newblock Spatio-temporal person retrieval via natural language queries.
\newblock In \emph{ICCV}, 2017.

\bibitem[Yang et~al.(2022)Yang, Miech, Sivic, Laptev, and Schmid]{yang2022tubedetr}
Antoine Yang, Antoine Miech, Josef Sivic, Ivan Laptev, and Cordelia Schmid.
\newblock Tubedetr: Spatio-temporal video grounding with transformers.
\newblock In \emph{CVPR}, 2022.

\bibitem[You et~al.(2023)You, Zhang, Gan, Du, Zhang, Wang, Cao, Chang, and Yang]{you2023ferret}
Haoxuan You, Haotian Zhang, Zhe Gan, Xianzhi Du, Bowen Zhang, Zirui Wang, Liangliang Cao, Shih-Fu Chang, and Yinfei Yang.
\newblock Ferret: Refer and ground anything anywhere at any granularity.
\newblock \emph{arXiv preprint arXiv:2310.07704}, 2023.

\bibitem[Yu et~al.(2021)Yu, Wang, Hu, Luo, and Li]{yu2022yu2rd}
Yi Yu, Xinying Wang, Wei Hu, Xun Luo, and Cheng Li.
\newblock 2rd place solutions in the hc-stvg track of person in context challenge 2021.
\newblock \emph{arXiv preprint arXiv:2106.07166}, 2021.

\bibitem[Zhang et~al.(2022)Zhang, Li, Liu, Zhang, Su, Zhu, Ni, and Shum]{zhang2022dino}
Hao Zhang, Feng Li, Shilong Liu, Lei Zhang, Hang Su, Jun Zhu, Lionel~M. Ni, and Heung-Yeung Shum.
\newblock Dino: Detr with improved denoising anchor boxes for end-to-end object detection.
\newblock \emph{arXiv preprint arXiv:2203.03605}, 2022.

\bibitem[Zhang et~al.(2020)Zhang, Zhao, Zhao, Wang, Liu, and Gao]{zhang2020stgrnVIDSTG}
Zhu Zhang, Zhou Zhao, Yang Zhao, Qi Wang, Huasheng Liu, and Lianli Gao.
\newblock Where does it exist: Spatio-temporal video grounding for multi-form sentences.
\newblock In \emph{CVPR}, 2020.

\bibitem[Zhang et~al.(2021)Zhang, Zhao, Lin, Huai, and Yuan]{zhang2022omrn}
Zhu Zhang, Zhou Zhao, Zhijie Lin, Baoxing Huai, and Jing Yuan.
\newblock Object-aware multi-branch relation networks for spatio-temporal video grounding.
\newblock In \emph{IJCAI}, 2021.

\bibitem[Zhou et~al.(2018)Zhou, Louis, and Corso]{zhou2018youcook2}
Luowei Zhou, Nathan Louis, and Jason~J Corso.
\newblock Weakly-supervised video object grounding from text by loss weighting and object interaction.
\newblock In \emph{BMVC}, 2018.

\bibitem[Zhu et~al.(2021)Zhu, Su, Lu, Li, Wang, and Dai]{zhu2021deformable}
Xizhou Zhu, Weijie Su, Lewei Lu, Bin Li, Xiaogang Wang, and Jifeng Dai.
\newblock Deformable detr: Deformable transformers for end-to-end object detection.
\newblock In \emph{ICLR}, 2021.

\end{thebibliography}
}

\end{document}